\documentclass[10pt,twocolumn,letterpaper]{article}

\usepackage[margin=0.95in]{geometry}
\usepackage{times}
\usepackage{graphicx}
\usepackage{amsmath}
\usepackage{amssymb}
\usepackage{booktabs}
\usepackage{multirow}
\usepackage{array}
\usepackage{xcolor}
\usepackage{hyperref}
\usepackage{enumitem}
\usepackage[T1]{fontenc}
\usepackage[utf8]{inputenc}
\usepackage{microtype}
\usepackage{arydshln}
\usepackage{titlesec}

\titleformat{\section}{\normalfont\large\bfseries}{\thesection.}{0.5em}{}
\titleformat{\subsection}{\normalfont\normalsize\bfseries}{\thesubsection}{0.5em}{}
\titlespacing*{\section}{0pt}{8pt plus 2pt minus 2pt}{4pt plus 1pt minus 1pt}
\titlespacing*{\subsection}{0pt}{6pt plus 2pt minus 1pt}{3pt plus 1pt minus 1pt}
\hypersetup{
    colorlinks=true,
    linkcolor=blue!60!black,
    citecolor=blue!60!black,
    urlcolor=blue!70!black
}

\newcommand{\etal}{\emph{et~al.}}
\newcommand{\pp}{\,pp}
\newcommand{\smol}{\textsc{SmolVLM2-500M}}
\newcommand{\qwen}{\textsc{Qwen2.5-VL-7B}}

\begin{document}

\title{Edge Reliability Gap in Vision-Language Models:\\ Quantifying Failure Modes of Compressed VLMs Under Visual Corruption}

\author{Mehmet Kaan Erol\\
{\small Marmara University, Institute of Pure and Applied Sciences}\\
{\small \texttt{kaanerol@marun.edu.tr}}\\
{\small \url{https://github.com/knrl/quantifying-vlms}}\\
}

\maketitle

\begin{abstract}
The rapid compression of large vision-language models (VLMs) for edge deployment raises an underexplored question: \emph{do compact models fail differently, not merely more often?} This study compares a 7-billion-parameter quantised VLM (Qwen2.5-VL-7B, 4-bit NF4) against a 500-million-parameter FP16 model (SmolVLM2-500M) across 4{,}000 samples from VQAv2 and COCO Captions on identical hardware (single NVIDIA RTX 5090). A three-category \textbf{error taxonomy} (Object Blindness, Semantic Drift, Prior Bias) is applied as a diagnostic framework; a text-only GPT-4o judge reveals Semantic Drift~(B) as the dominant failure mode on VQAv2 and on COCO for Qwen, with a mixed Object Blindness / Semantic Drift profile for SmolVLM2 on COCO; Prior Bias~(C) is present on VQAv2 but absent on COCO for both models. \textbf{Confidence calibration} is measured via Expected Calibration Error (ECE) using geometric mean token probability (\S\ref{sec:ece_method}); compositional reasoning is probed with structured \textbf{negation probes} across four templates; and a \textbf{blur robustness} experiment completes the evaluation. For this model pair, the compact model exhibits a qualitatively distinct failure signature: a $12.5\pp$ larger negation collapse ($-33.2\pp$ vs.\ $-20.8\pp$; Wald 95\% CI $[8.2, 16.8]\pp$, $p < 10^{-8}$), driven almost entirely by COCO (per-dataset gap $20.5\pp$, $p < 10^{-16}$) while the VQAv2 gap is not statistically significant ($4.5\pp$, $p{=}0.19$). The most discriminating template is \texttt{false\_yn}: \smol{} responds ``Yes'' (incorrectly claiming a depicted object is absent) on 100\% of COCO trials vs.\ 14\% for \qwen{}. Asymmetric dataset-dependent miscalibration and a blur experiment with two controlled ablations complete the analysis. The fully reproducible pipeline is released for systematic safety auditing of compressed VLMs prior to edge deployment.
\end{abstract}

\section{Introduction}
\label{sec:intro}

The proliferation of multimodal AI at the network edge---in smartphones, autonomous vehicles, robotics, and IoT devices---has created enormous pressure to compress large vision-language models~\cite{laurencon2024smolvlm,chu2023mobilevlm,zhu2024llavaphi}. Techniques including quantisation~\cite{dettmers2023qlora,frantar2022gptq,lin2024awq}, pruning~\cite{fang2023depgraph,shang2024llavaprumerge}, and architecture miniaturisation~\cite{laurencon2024smolvlm,abdin2024phi3} have produced models that run comfortably within 1--5\,GiB VRAM budgets. The standard evaluation narrative is straightforward: smaller or more quantised models achieve lower aggregate accuracy. However, \textbf{aggregate accuracy conceals the structure of failure}.

Consider two models with similar accuracy on a VQA benchmark. One consistently misses rare object categories (a perceptual failure); the other hallucinates plausible but incorrect attributes with high confidence (a linguistic failure). These models carry fundamentally different deployment risks. An autonomous vehicle assistant that says ``I don't see a stop sign'' (uncertainty-aware) is preferable to one that confidently asserts ``there is no stop sign'' when one is clearly present (overconfident hallucination)~\cite{guan2024hallusionbench,li2023pope}. For safety-critical edge applications, the \emph{type} of error matters as much as its \emph{rate}.

Despite extensive work on VLM compression~\cite{kundu2025lvlmcompressbench,dettmers2023qlora,frantar2022gptq,lin2024awq,xiao2023smoothquant} and hallucination benchmarking~\cite{li2023pope,rohrbach2018chair,guan2024hallusionbench,lovenia2023negatedhallu}, no prior study has jointly measured \emph{error taxonomy}, \emph{calibration quality}, and \emph{negation robustness} across size regimes for compressed VLMs on identical hardware. This gap is instantiated for one representative compact-vs.-large VLM pair; the primary contribution is the evaluation methodology and diagnostic framework:

\begin{itemize}[leftmargin=*,nosep]
    \item[\textbf{C1.}] A three-category error taxonomy---\emph{Object Blindness}, \emph{Semantic Drift}, \emph{Prior Bias}---as a \textbf{reusable diagnostic framework} for VLM failures, with two auxiliary categories (Spatial/Relational Error~D, Other~E) for residual cases (\S\ref{sec:taxonomy}). A heuristic classifier provides a zero-cost reproducibility baseline; a text-only GPT-4o judge ($n{=}400$ API calls, no image access) provides the authoritative labels in Table~\ref{tab:taxonomy} (prompt in Appendix~\ref{app:judge_prompt}). A vision-capable re-judge remains an open extension.
    
    \item[\textbf{C2.}] The edge reliability gap for this model pair: \smol{} achieves 47.8\% VQAv2 / 92.2\% COCO accuracy vs.\ \qwen{} at 55.6\% / 91.0\%; ECE of 0.329 vs.\ 0.265 (average); negation drop of $-33.2\pp$ vs.\ $-20.8\pp$. Because the two models differ simultaneously in scale, precision, and architecture, these differences characterise the specific pairing rather than isolating any single factor (\S\ref{sec:confound}). A blur experiment ($n{=}100$) finds equal degradation for both models ($\Delta{=}3\pp$ each, $\rho{=}1.0$, McNemar $p{=}0.68$, not significant); controlled ablations show NF4 quantisation adds $3\pp$ blur sensitivity and scale within the Qwen architecture confers robustness ($\rho_{\text{arch}}{=}0.75$ for SmolVLM2 vs.\ Qwen-3B FP16).
    
    \item[\textbf{C3.}] A fully reproducible evaluation pipeline encompassing all experiments, dataset streaming, and figure generation, enabling systematic safety auditing of VLMs prior to edge deployment.
\end{itemize}

\section{Related Work}
\label{sec:related}

\subsection{VLM Compression for Edge Deployment}

Model compression for vision-language tasks has been approached through three complementary strategies. \textbf{Weight quantisation} methods including GPTQ~\cite{frantar2022gptq}, bitsandbytes NF4~\cite{dettmers2023qlora}, AWQ~\cite{lin2024awq}, and SmoothQuant~\cite{xiao2023smoothquant} dramatically reduce memory requirements but are primarily evaluated on language benchmarks, leaving vision-language grounding quality under deployment conditions under-studied. \textbf{Structured pruning} approaches such as DepGraph~\cite{fang2023depgraph}, SlimVLM~\cite{dai2024slimvlm}, and LLaVA-PruMerge~\cite{shang2024llavaprumerge} reduce model footprint but reveal that visual projection layers are disproportionately sensitive. \textbf{Architecture miniaturisation} via MobileVLM~\cite{chu2023mobilevlm}, LLaVA-Phi~\cite{zhu2024llavaphi}, Phi-3.5-Vision~\cite{abdin2024phi3}, and the SmolVLM series~\cite{laurencon2024smolvlm} pushes the parameter frontier below 1B, demonstrating non-trivial VQA performance at significant accuracy cost on fine-grained tasks.

Recent systematic benchmarks reveal that compression effects extend beyond accuracy. LVLM-Compress-Bench~\cite{kundu2025lvlmcompressbench} shows that weight and KV-cache quantisation produce capability-specific trade-offs across recognition, spatial reasoning, and hallucination. Bouguerra \etal~\cite{bouguerra2025quantclip} demonstrate that CLIP quantisation always impacts reliability---altering calibration, OOD detection, and shift robustness---with effects depending on pre-training data and quantisation recipe. Bhatnagar \etal~\cite{bhatnagar2025luq} show that multimodal tokens exhibit higher variance than text tokens, making some layers highly sensitive to sub-4-bit quantisation. Crucially, across all these lines of work, the \emph{structural characteristics} of remaining errors are not systematically reported.

\subsection{Hallucination in Vision-Language Models}

Hallucination benchmarks have established important categories of VLM failure. POPE~\cite{li2023pope} measures object existence hallucination via binary polling. CHAIR~\cite{rohrbach2018chair} quantifies hallucinated objects in captions. HallusionBench~\cite{guan2024hallusionbench} constructs adversarial visual illusions to disentangle language-prior hallucination from visual perception errors. NegatedHallu~\cite{lovenia2023negatedhallu} and HallE-Control~\cite{zhai2023hallecontrol} study hallucination at the small-model end, finding that lightweight models exhibit qualitatively different hallucination profiles: more frequent Prior Bias and fewer Semantic Drift errors.

Hallucination mitigation has been approached through RLHF~\cite{sun2023llava_rlhf}, retrospective revision~\cite{zhou2023lure}, visual contrastive decoding~\cite{leng2024vcd}, and attention penalty methods~\cite{huang2024opera}. Importantly, all of these interventions are developed and evaluated for models with $\geq$7B parameters. Their effectiveness at the 0.5B scale---where language prior dominance is likely stronger---remains unknown. Comprehensive surveys~\cite{huang2023survey_hallucination,kou2024hallucination_survey} frame hallucination as a lifecycle-wide failure mode and emphasise structured uncertainty estimation as a diagnostic backbone.

\subsection{Robustness Under Visual Corruption}

ImageNet-C~\cite{hendrycks2019imagenetc} established synthetic corruptions as a standard robustness benchmark. For VLMs, MMRobustBench~\cite{li2024mmrobustbench} introduces 14 visual corruptions across seven benchmarks, finding that robustness and accuracy are weakly correlated. Tu \etal~\cite{tu2024cliprobust} evaluate 84 CLIP models under ten corruption types, finding that encoder architecture strongly affects blur sensitivity and that fine-tuning can reduce beneficial shape biases. Bao \etal~\cite{bao2025mint} show that common corruptions induce ``embedding variance collapse'' in CLIP encoders---intra-class and inter-class variances shrink with corruption severity. Ramos \etal~\cite{ramos2025processing} demonstrate that processing and acquisition shifts are systematically encoded in visual features and can dominate semantic predictions.

Geirhos \etal~\cite{geirhos2019texture} show that texture bias in vision models makes them particularly sensitive to blur, which suppresses high-frequency detail. For VLMs with dynamic resolution systems like Qwen2.5-VL, multi-scale tiling may partially compensate for blur degradation, whereas fixed-resolution encoders like SigLIP-400M may lose high-frequency information more readily. This architectural difference---among several confounded factors including model scale and precision---motivates our blur robustness hypothesis (\S\ref{sec:robustness}).

\subsection{Confidence Calibration}

A classifier is \emph{calibrated} if its predicted confidence matches its empirical accuracy~\cite{guo2017calibration}. Expected Calibration Error (ECE)~\cite{naeini2015ece} is the canonical scalar summary. Modern deep networks are systematically overconfident~\cite{guo2017calibration}, and RLHF fine-tuning consistently degrades calibration~\cite{xiong2024llm_calibration}. For VLMs, calibration is substantially less studied. Xuan \etal~\cite{xuan2025verbalized_calibration} show systematic miscalibration across VLM types and develop Visual Confidence-Aware Prompting. Zhao \etal~\cite{zhao2025csp} find that VLMs output near-100\% verbalized confidence on hallucinated objects. Groot \& Valdenegro-Toro~\cite{groot2024overconfidence} report persistent overconfidence in GPT-4V and Gemini Pro Vision. The interaction between quantisation and calibration is largely unexplored; our study provides the first ECE comparison between a 4-bit quantised 7B model and an FP16 0.5B model.

\subsection{Negation and Compositional Reasoning}

Negation remains challenging for neural language models. Ettinger~\cite{ettinger2020bert} demonstrates that BERT-family models fail negation tests via shallow lexical matching. NegatedHallu~\cite{lovenia2023negatedhallu} finds that LLaVA-1.5-7B incorrectly agrees with false negative visual statements in over 40\% of cases. NegVQA~\cite{zhao2023negvqa} constructs naturally negated VQA questions, finding a consistent negation performance gap that is larger for smaller models. Mechanistic accounts~\cite{yu2023negation,ravichander2022condaqa} suggest that attention-based models may struggle to propagate negation scope effectively---the token ``not'' modifies semantics locally, but attention patterns can dilute this signal. Empirically, this effect appears to be associated with smaller LLM backbones~\cite{zhao2023negvqa}, though whether the underlying mechanism is insufficient attention capacity, inadequate training data coverage of negated constructions, or some other factor remains an open question.

\section{Evaluation Framework}
\label{sec:framework}

\subsection{Models Under Study}

Two models spanning a representative size-vs-efficiency frontier are selected, both deployable on a single consumer GPU (Table~\ref{tab:models}).

\begin{table}[t]
\centering
\caption{Models under study. Both are loaded simultaneously on a single NVIDIA RTX 5090 (32\,GiB VRAM). Combined footprint $\approx$5.5--6\,GiB.}
\label{tab:models}
\small
\begin{tabular}{@{}lcc@{}}
\toprule
\textbf{Property} & \textbf{\qwen} & \textbf{\smol} \\
\midrule
Provider & Alibaba / Qwen & HuggingFace \\
Parameters & $\sim$7.6B & $\sim$0.5B \\
Precision & NF4 4-bit & FP16 \\
VRAM & $\sim$4.5\,GiB & $\sim$1.0\,GiB \\
Vision encoder & Qwen2.5-VL & SigLIP-400M \\
LLM backbone & Qwen2.5-7B & SmolLM2-135M \\
\bottomrule
\end{tabular}
\end{table}

\textbf{\qwen{}} is a 7B-parameter model from Alibaba~\cite{bai2025qwen25vl}, quantised to 4-bit NF4 via bitsandbytes with double quantisation and \texttt{float16} compute dtype. Its native dynamic resolution system tiles high-resolution images into variable-length visual token sequences, providing multi-scale representation. \textbf{\smol{}} is a 500M-parameter model from HuggingFace~\cite{laurencon2024smolvlm}, running in native FP16. It uses SigLIP-400M as its vision encoder and SmolLM2-135M as its language backbone.

\textbf{Design note (three-variable confound).} This comparison simultaneously varies scale (7B vs.\ 0.5B), numerical precision (NF4 vs.\ FP16), and architecture (Qwen2.5-VL vs.\ SmolVLM2). All reported differences are therefore \emph{associative} rather than causally attributed to any single factor. Two controlled ablations (\S\ref{sec:confound}) partially disentangle these factors by adding Qwen2.5-VL-7B in native FP16 (isolates precision) and Qwen2.5-VL-3B in FP16 (isolates scale within architecture), forming a partial $2{\times}2$ factorial over precision and scale within the Qwen2.5-VL architecture family.

\subsection{Datasets}

Two established benchmarks are used (Table~\ref{tab:datasets}), streamed via HuggingFace \texttt{datasets} with \texttt{streaming=True}. Corrupted or zero-size images are skipped and logged.

\begin{table}[t]
\centering
\caption{Evaluation datasets. Both are streamed; results are continuously flushed with batch-level checkpointing.}
\label{tab:datasets}
\small
\begin{tabular}{@{}lccc@{}}
\toprule
\textbf{Dataset} & \textbf{Split} & \textbf{Samples} & \textbf{Task} \\
\midrule
VQAv2~\cite{goyal2017vqav2} & validation & 2{,}000 & Open VQA \\
COCO Captions & val & 2{,}000 & Captioning \\
\bottomrule
\end{tabular}
\end{table}

\textbf{Correctness metric.} For VQAv2, a prediction is correct if the normalised ground-truth string is a substring of the normalised prediction or vice versa (soft match), approximating the official 10-annotator VQA metric. For COCO Captions, correctness requires keyword overlap: the prediction must contain at least one content word (length $\geq$3) from the ground-truth caption. This is a deliberately coarse proxy whose weakness has \textbf{concrete downstream consequences}. Because matching a single common noun suffices, captions with severe semantic errors may still register as ``correct'', inflating both models' COCO accuracy and compressing the accuracy gap. The near-parity between \smol{} (92.2\%) and \qwen{} (91.0\%) on COCO is therefore \textbf{not interpretable as evidence of equivalent captioning quality}---standard captioning metrics (CIDEr, METEOR, BLEU-4) would likely reveal a larger quality gap. Including even one standard metric (CIDEr is trivially available via \texttt{pycocoevalcap}) would ground the COCO comparison; this is left for future work and is a clear limitation.

\textbf{Implications for negation experiments.} The COCO ``both-correct'' set---on which the negation probes are computed---is defined using this coarse keyword-overlap criterion. If any rows in the both-correct set are false positives (semantically incorrect captions that happen to share a content word with the reference), the negation probes for those rows operate on an unreliable baseline. Because the keyword-overlap criterion is \emph{permissive} rather than strict, false positives are the primary risk: the both-correct set may include rows where one or both models already produced poor captions. The 90.0\% both-correct rate for COCO is consistent with this concern (a stricter metric would likely yield a lower rate). The negation gap itself---which measures relative change from baseline---is less affected than the absolute accuracy levels, but readers should bear the coarse metric in mind when interpreting COCO negation results.

\subsection{Error Taxonomy}
\label{sec:taxonomy}

Three primary error categories target structurally distinct failure modes, plus two auxiliary categories:

\begin{itemize}[leftmargin=*,nosep]
    \item \textbf{Object Blindness (A):} Model fails to identify a salient object that is visually prominent. The answer references an irrelevant object or claims nothing is visible.
    \item \textbf{Semantic Drift (B):} Model identifies the correct object category but produces a semantically inconsistent description---wrong colour, count, action, or spatial relation.
    \item \textbf{Prior Bias (C):} Model generates a plausible response consistent with statistical priors about the scene, independent of actual image content.
    \item \textbf{Spatial/Relational Error (D):} Model produces an incorrect spatial relation, object layout, or count. This category captures failures that involve quantitative or positional reasoning rather than object identity or attribute errors. Category~D failures are excluded from the three-way A/B/C percentage distribution but are reported as raw counts (Table~\ref{tab:taxonomy}).
    \item \textbf{Other (E):} Failure does not clearly fit A--D. Excluded from the A/B/C distribution; raw counts reported.
\end{itemize}

Categories A--C are mutually exclusive; each failure receives the label of its \textbf{most proximal cause}---the failure mode that, if corrected alone, would make the prediction correct. Categories D and E capture residual cases and are excluded from the A/B/C percentages in Table~\ref{tab:taxonomy}. The heuristic applies D \emph{before} B: any question containing spatial or counting keywords is assigned D regardless of token overlap, so D counts may include cases that would otherwise qualify as B. The GPT-4o judge prompt (Appendix~\ref{app:judge_prompt}) includes all five categories; the main text focuses on A--C as the primary failure modes of interest. Appendix~\ref{app:qualitative} Example~3 illustrates a Category~D case.

\subsection{Heuristic Taxonomy Classifier}
\label{sec:judge}

Taxonomy labels are assigned by a rule-based heuristic classifier, which applies a deterministic decision tree to the question, ground-truth, and prediction strings: Spatial/Relational~(D) if the question contains spatial or count keywords; Object Blindness~(A) if negation words appear or no ground-truth tokens overlap the prediction; Semantic Drift~(B) if partial ground-truth token overlap is present; Prior Bias~(C) if the prediction is verbally fluent but shares no tokens with the ground truth; Other~(E) otherwise. The classifier has no image access. A text-only GPT-4o-as-judge~\cite{zheng2023mtbench,chiang2023llmjudge} was subsequently executed as the primary taxonomy judge (March 2026; $n{=}400$ API calls, text only, no image access); results appear in Table~\ref{tab:taxonomy} and \S\ref{sec:taxonomy_results}. The GPT-4o prompt is preserved in Appendix~\ref{app:judge_prompt}. The heuristic remains a zero-cost reproducibility fallback; note that both judges operate on text only (question, ground-truth, prediction) without image access---a vision-capable re-judge remains an open extension.

\textbf{Implementation note.} SmolVLM2 outputs retain the chat-template prefix ``\texttt{Assistant:}'' after decoding. A \texttt{clean\_prediction()} helper strips this prefix before taxonomy classification to prevent spurious token overlap from corrupting the A/B/C decision logic. The correctness metrics are unaffected because their normalisation already removes such tokens.

A \emph{failure-profile concordance} score is computed using Cohen's $\kappa$ formula~\cite{cohen1960kappa} between taxonomy labels for each model's failures. Because the two failure sets consist of different images, this is \textbf{not} standard inter-rater $\kappa$; it measures whether the taxonomy categories are used with similar relative frequency across models---a descriptive comparison, not a reliability estimate. True inter-rater $\kappa$ (requiring shared cases) is left as future work.

\subsection{Gaussian Blur Perturbation}
\label{sec:blur_method}

Visual robustness is measured by re-running both models on Gaussian-blurred versions of a 100-sample subset where both models were originally correct (the \emph{baseline-correct} set). Blur is applied (kernel $5{\times}5$, $\sigma{=}2.0$), corresponding approximately to severity level~2 on the ImageNet-C scale~\cite{hendrycks2019imagenetc}. This severity was chosen to produce non-trivial but not catastrophic degradation---the inflection point where encoder resolution differences are most diagnostic.

The relative robustness ratio $\rho$ quantifies differential blur sensitivity:
\begin{equation}
\rho = \frac{\Delta_{\text{Smol}}}{\Delta_{\text{Qwen}}}\,,
\label{eq:rho}
\end{equation}
where $\Delta_{\text{model}} = \text{Acc}_{\text{clean}} - \text{Acc}_{\text{blurred}}$. $\rho > 1$ indicates that \smol{} is proportionally more sensitive to blur.

\subsection{Confidence Calibration}
\label{sec:ece_method}

Per-token log-probabilities are aggregated into a sequence-level confidence score via geometric mean token probability:
\begin{equation}
\text{conf}(s) = \exp\!\left(\frac{1}{|s|}\sum_{t=1}^{|s|} \log p(s_t \mid s_{<t}, x)\right)\,,
\label{eq:conf}
\end{equation}
where $s$ is the generated token sequence and $x$ is the multimodal input. Predictions are binned into $M{=}10$ equal-width confidence bins over $[0, 1]$. Expected Calibration Error~\cite{naeini2015ece} is:
\begin{equation}
\text{ECE} = \sum_{m=1}^{M} \frac{|B_m|}{N} \left| \text{acc}(B_m) - \overline{\text{conf}}(B_m) \right|\,,
\label{eq:ece}
\end{equation}
where $|B_m|$ is the bin population, $\text{acc}(B_m)$ is fraction correct in bin $m$, and $\overline{\text{conf}}(B_m)$ is the arithmetic mean of per-sequence geometric-mean confidences within bin~$m$. This two-level aggregation---geometric mean within sequences, arithmetic mean across sequences in a bin---is consistent with standard LLM calibration practice~\cite{kadavath2022know}.

\textbf{Proxy validity caveats.} Three limitations apply. First, for yes/no VQA questions, sequence-level log-probabilities conflate answer confidence with generation fluency; a first-token proxy would be cleaner. Second, variable-length COCO captions accumulate low-probability tokens, artificially lowering the geometric mean and likely contributing to \smol{}'s COCO underconfidence (ECE~$=0.431$). Third, Kadavath \etal~\cite{kadavath2022know} elicit explicit ``P(True)'' confidence rather than extracting it from generation log-probabilities; the approaches are methodologically distinct. Despite these caveats, token log-probability geometric mean is the standard proxy for auto-regressive models, and the most anomalous finding---\qwen{}'s constant $\approx$0.999 confidence on VQAv2---is robust to any reasonable proxy choice.

\subsection{Negation Stress Tests}
\label{sec:negation_method}

Four negation operator templates are applied to up to 100 \emph{both-correct} rows per dataset (Table~\ref{tab:negation_templates}), yielding up to 800 judgements per model per dataset. Baseline accuracy is 100\% by construction, making the negation drop a clean measure of compositional sensitivity.

\begin{table}[t]
\centering
\caption{Negation probe templates and correctness criteria. For \texttt{false\_yn}, the placeholder \texttt{\{answer\}} is the model's \emph{original correct answer}---an object that \textbf{is} depicted. The question therefore asks whether something present is absent; the semantically correct reply is \textbf{``no''}. A model that says ``yes'' is hallucinating (incorrectly claiming the depicted object is absent).}
\label{tab:negation_templates}
\small
\begin{tabular}{@{}lp{3.8cm}@{}}
\toprule
\textbf{Template} & \textbf{Correctness criterion} \\
\midrule
\texttt{is\_not} & Does \emph{not} contain original answer \\
\texttt{absent} & Does \emph{not} contain original answer \\
\texttt{false\_yn} & Contains ``no'' in the first 20 normalised characters (rejects false premise)\footnotemark \\
\texttt{counter} & Contains a non-trivial word $\neq$ original \\
\bottomrule
\end{tabular}
\end{table}
\footnotetext{Implemented as \texttt{"no" in resp[:20]} in \texttt{negation\_probes.py}: the response is lowercased and stripped, then the first 20 characters are checked for the substring ``no''. This window is narrow enough to capture the answer token while avoiding false positives from explanatory text later in the response.}

A model that collapses ``no dog'' into ``dog'' at the embedding level will answer as if the negation were absent. The \texttt{false\_yn} template is the most discriminating: it asserts a false premise about a depicted object being absent, and the correct response is ``No''. A model responding ``Yes'' is hallucinating.

\section{Experiments and Results}
\label{sec:experiments}

All experiments run on a single NVIDIA RTX 5090 (32\,GiB VRAM) with deterministic seeding (\texttt{SEED=42}), \texttt{torch.backends.cudnn.deterministic=True}, and \texttt{benchmark=False}. The pipeline consists of six sequential phases: sanity check, batch inference, blur robustness, ECE calibration, negation probes, and LLM-as-judge taxonomy validation.

\subsection{Baseline Accuracy}
\label{sec:baseline}

Table~\ref{tab:baseline} presents clean-image accuracy on 2{,}000 samples per dataset. \qwen{} leads \smol{} by $+7.8\pp$ on VQAv2 but trails by $-1.2\pp$ on COCO Captions. The COCO near-parity should be interpreted with caution: the single-keyword overlap metric used for COCO correctness (\S\ref{sec:framework}) is deliberately coarse, and near-parity under this metric is not evidence of equivalent captioning quality---standard metrics (CIDEr, METEOR, BLEU-4) would likely reveal a larger quality gap. VQAv2 demands fine-grained object recognition and attribute reasoning where the larger model's richer representations are associated with higher accuracy.

\begin{table}[t]
\centering
\caption{Clean-image accuracy ($n{=}2{,}000$ per dataset).}
\label{tab:baseline}
\small
\begin{tabular}{@{}lccc@{}}
\toprule
\textbf{Model} & \textbf{VQAv2} & \textbf{COCO} & \textbf{Combined} \\
\midrule
\smol{} (FP16) & 47.8\% & 92.2\% & 70.0\% \\
\qwen{} (NF4) & 55.6\% & 91.0\% & 73.3\% \\
\midrule
$\Delta$ (Qwen$-$Smol) & +7.8\pp & $-$1.2\pp & +3.3\pp \\
\bottomrule
\end{tabular}
\end{table}

\subsection{Error Taxonomy Distribution}
\label{sec:taxonomy_results}

A text-only GPT-4o semantic judge assigned taxonomy labels to $n{=}100$ failures per model $\times$ dataset ($400$ labels total). The judge receives the question, ground-truth answer, and model prediction as text and returns a structured JSON label (category A--E plus free-text reason); no image is sent to the judge. A rule-based heuristic is also available as a zero-cost reproducibility fallback. As with all model comparisons in this study, the three-variable confound applies (\S\ref{sec:confound}).

\textbf{VQAv2 taxonomy.} Table~\ref{tab:taxonomy} shows the distribution for VQAv2 failures. Semantic Drift~(B) dominates for both models (SmolVLM2: 69.4\%, Qwen NF4: 51.9\% of A--C labels), consistent with both models identifying the correct object category but producing wrong attributes (colour, count, or relation). Qwen NF4 shows more Prior Bias~(C) (20.4\% vs.\ 4.2\%), suggesting the larger model generates more plausible-but-wrong predictions anchored to scene priors rather than specific image content. Spatial/Relational Errors~(D) are 19 (SmolVLM2) and 24 (Qwen), reflecting VQAv2's spatial and counting questions; these are excluded from the A--C percentage. Both models always produce text in failure cases (0\% empty-output rate).

\textbf{COCO Captions taxonomy.} The GPT-4o judge breaks the 100\% Object Blindness artefact that the keyword-mismatch heuristic produced for COCO. Semantic Drift~(B) is dominant for Qwen NF4 (82.8\% of A--C labels), and the two failure categories are roughly split for SmolVLM2 (Object Blindness~A 46.6\%, Semantic Drift~B 53.4\%). Prior Bias~(C) is 0\% for both models on COCO, consistent with captions being open-ended descriptions where statistical scene priors are harder to confuse with specific errors. Spatial Error~(D) is rare (SmolVLM2:~8, Qwen:~3), as expected for free-form captioning. The markedly different COCO failure profiles between the two models (Qwen: Semantic Drift dominant; SmolVLM2: split A/B) suggest that the larger model's stronger language backbone produces caption text that semantically misrepresents scene content, while the compact model more often omits key objects entirely.

\begin{table*}[t]
\centering
\caption{Error taxonomy distribution from GPT-4o text-only judge (\% of A--C labelled failures; $n{=}100$ per model $\times$ dataset, seed 42). D and E reported as raw counts; both are excluded from the A--C percentage denominator.}
\label{tab:taxonomy}
\small
\begin{tabular*}{\textwidth}{@{\extracolsep{\fill}}llcc@{}}
\toprule
\textbf{Dataset} & \textbf{Error Type} & \textbf{\smol} & \textbf{\qwen} \\
\midrule
\multirow{5}{*}{VQAv2} & Object Blindness (A) & 26.4\% & 27.8\% \\
 & Semantic Drift (B) & 69.4\% & 51.9\% \\
 & Prior Bias (C) & 4.2\% & 20.4\% \\
\hdashline
 & Spatial Error (D) & (19) & (24) \\
 & Other (E) & (9) & (22) \\
\midrule
\multirow{5}{*}{COCO} & Object Blindness (A) & 46.6\% & 17.2\% \\
 & Semantic Drift (B) & 53.4\% & 82.8\% \\
 & Prior Bias (C) & 0.0\% & 0.0\% \\
\hdashline
 & Spatial Error (D) & (8) & (3) \\
 & Other (E) & (4) & (4) \\
\bottomrule
\end{tabular*}
\end{table*}

\begin{figure*}[t]
\centering
\includegraphics[width=\textwidth]{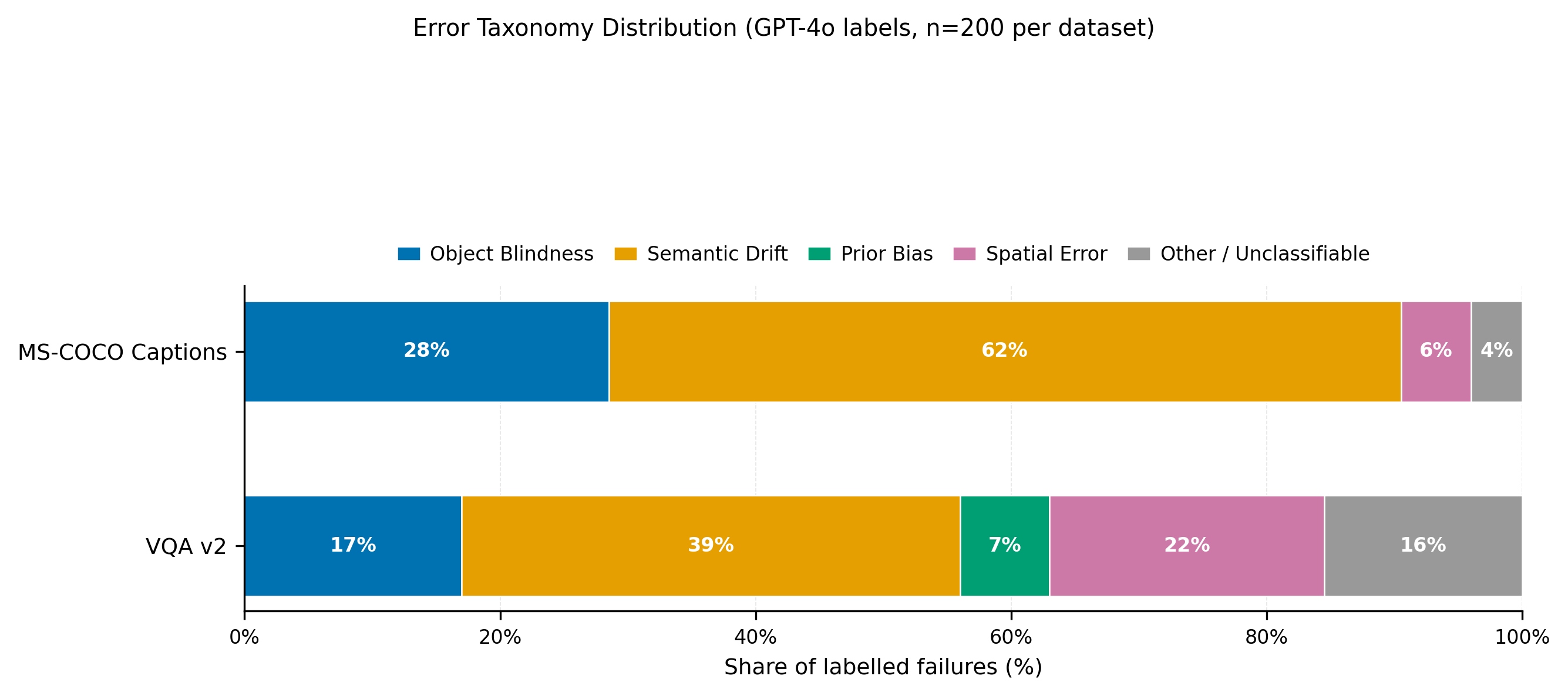}
\caption{GPT-4o text-only error-taxonomy distribution across VQAv2 and COCO Captions ($n{=}200$ failures per dataset; SmolVLM2 and Qwen NF4 averaged). VQAv2: Semantic Drift~(B) dominates (39.0\% combined), followed by Spatial Error~(D, 21.5\%); Object Blindness~(A) is 17.0\% and Prior Bias~(C) is 7.0\%. COCO: Semantic Drift~(B) dominates (62.0\% combined), with Object Blindness~(A) at 28.5\%; Prior Bias~(C) is 0\% for both models.}
\label{fig:taxonomy}
\end{figure*}

\textbf{Failure-profile concordance.} The failure-profile concordance score (Cohen's $\kappa$ formula applied across different failure sets; see \S\ref{sec:judge}) is $\kappa = 0.017$ on VQAv2 (slight), consistent with the two models exhibiting broadly similar failure distributions on VQAv2 but with \qwen{} showing more Prior Bias~(C) (20.4\%) than \smol{} (4.2\%). On COCO Captions, $\kappa = -0.038$ (less than chance), reflecting the divergent COCO failure profiles: \qwen{} concentrates Semantic Drift~(B, 82.8\%) while \smol{} is split between Object Blindness and Semantic Drift (46.6\%~A / 53.4\%~B). A proper inter-rater consistency estimate (test-retest double-labelling of a shared subset) is left as future work.

\begin{figure*}[t]
\centering
\includegraphics[width=\textwidth]{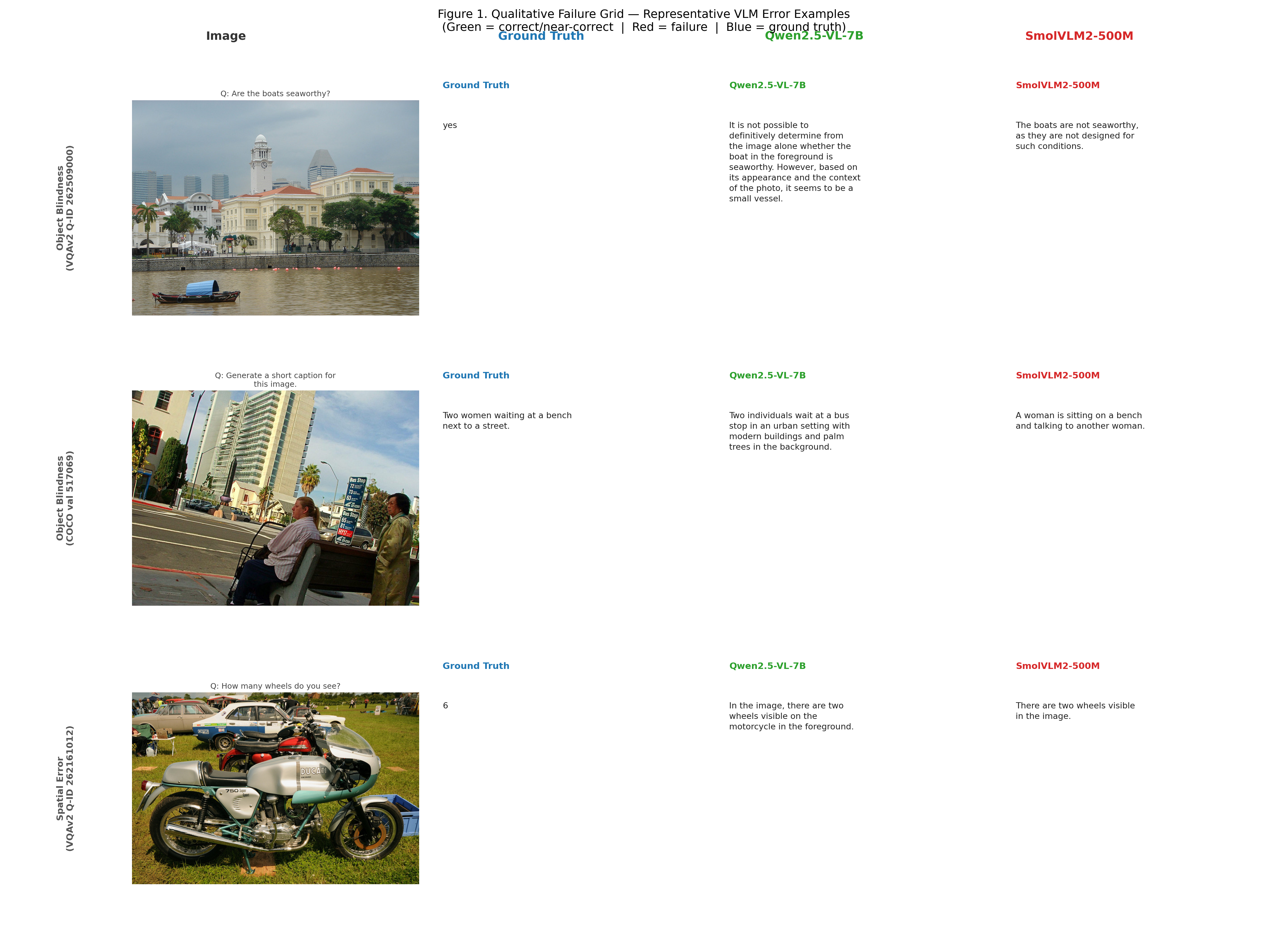}
\caption{Qualitative failure grid: three representative errors from \texttt{llm\_judge\_labels.json}. Columns show the input image, ground truth, \qwen{} output (green), and \smol{} output (red). Rows~1--2: Object Blindness on VQAv2 and COCO. Row~3: Spatial Error (VQAv2).}
\label{fig:failure_grid}
\end{figure*}

\subsection{Confidence Calibration}
\label{sec:ece_results}

Table~\ref{tab:ece} presents ECE results across both models and datasets. The three-variable confound (\S\ref{sec:confound}) applies: the ECE differences below are a joint function of scale, precision, and architecture. In particular, Qwen's confidence saturation ($p\approx 0.999$ for all predictions) may be partly an artefact of aggregating token-level log-probabilities over short VQA answers in a closed-answer setting---the softmax probability of the dominant token approaches 1.0 when the model's instruction-tuned distribution is sharply peaked, independent of genuine calibration quality. The finding is nonetheless operationally meaningful: regardless of its cause, a constant confidence function provides zero discriminative signal for deployment gating. The calibration landscape is asymmetric and dataset-dependent---a finding with important implications for confidence-based deployment gating.

\begin{table}[t]
\centering
\caption{Expected Calibration Error ($M{=}10$ bins, $n{=}2{,}000$ per dataset). Lower is better; a perfectly calibrated model has ECE~$= 0$.}
\label{tab:ece}
\small
\begin{tabular}{@{}lccc@{}}
\toprule
\textbf{Model} & \textbf{VQAv2} & \textbf{COCO} & \textbf{Average} \\
\midrule
\smol{} (FP16) & 0.228 & 0.431 & 0.329 \\
\qwen{} (NF4) & 0.443 & 0.087 & 0.265 \\
\bottomrule
\end{tabular}
\end{table}

\textbf{VQAv2 calibration.} \qwen{} exhibits the degenerate extreme of miscalibration on VQAv2: all 2{,}000 predictions fall in the top confidence bin ($\geq$0.9; mean confidence 0.999) while achieving only 55.6\% accuracy, yielding ECE $= 0.443$. This is \textbf{not routine overconfidence---it is a constant confidence function}. Any threshold set below 0.999 passes every prediction unconditionally (including all erroneous ones); any threshold at or above 0.999 rejects everything. \qwen{}'s token-level confidence carries zero discriminative signal for deployment gating on this task. \smol{} distributes predictions across multiple bins (bins 3--9) with moderate overconfidence (ECE $= 0.228$), meaning that while its confidences are imperfect, they carry some discriminative signal---a notable contrast with \qwen{}'s saturated output.

\textbf{COCO calibration.} The pattern reverses. \qwen{} places all 2{,}000 predictions in the top bin again, but with 91.1\% accuracy the miscalibration is much smaller (ECE $= 0.087$). \smol{} clusters predictions predominantly in bins 3--6 (confidence 0.37--0.63) while achieving 92.2\% accuracy, resulting in severe \emph{underconfidence} miscalibration (ECE $= 0.431$).

\begin{figure*}[t]
\centering
\includegraphics[width=\textwidth]{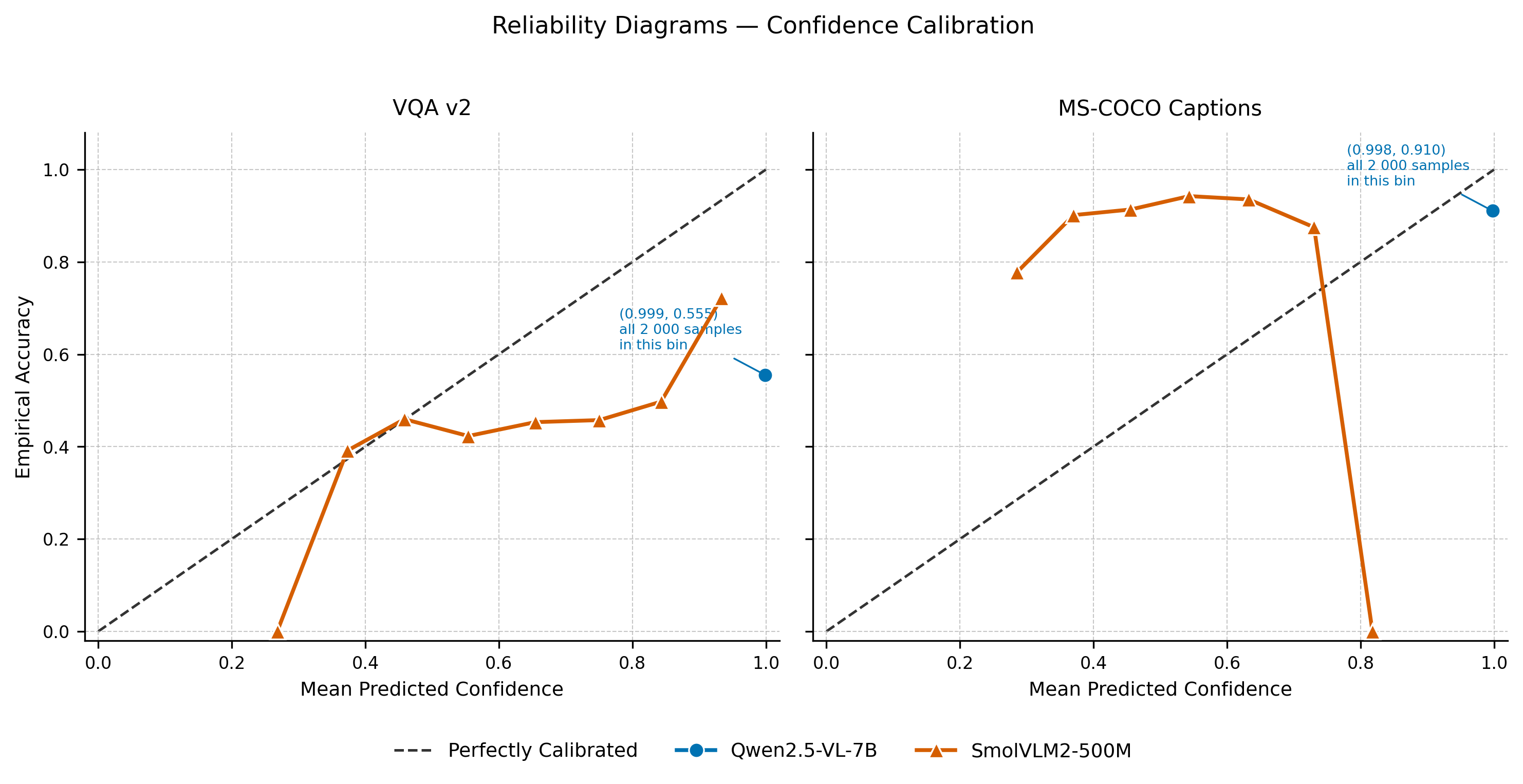}
\caption{Reliability diagrams for both models on VQAv2 (left) and COCO Captions (right). The dashed diagonal represents perfect calibration. \qwen{} (blue) on VQAv2 collapses its entire 2{,}000-prediction distribution to a \textbf{single point} at confidence $\approx$0.999, accuracy $\approx$0.556---annotated and isolated far to the right of the diagonal. This is the defining visual of confidence-function degeneration: the model issues maximum confidence unconditionally, regardless of correctness. \smol{} (red, dashed) distributes predictions across multiple bins: below the diagonal on VQAv2 (overconfident) and above it on COCO (underconfident).}
\label{fig:reliability}
\end{figure*}

\textbf{Implications.} \qwen{}'s saturated confidence on VQAv2 renders any confidence-based gate equivalent to no gate---every prediction is admitted at maximum certainty, including the 44.4\% that are wrong. Conversely, \smol{}'s COCO underconfidence would suppress many correct answers. No single threshold works across tasks for either model; task-adaptive calibration~\cite{guo2017calibration} is essential. Figure~\ref{fig:reliability} visualises this pattern.

\subsection{Robustness Under Gaussian Blur}
\label{sec:robustness}

Table~\ref{tab:blur} presents accuracy under Gaussian blur ($\sigma{=}2.0$) on the $n{=}100$ baseline-correct subset.\footnote{Both models answered correctly on all 100 images under clean conditions, giving a 100\% clean baseline by construction. 95\% bootstrap CIs (10{,}000 resamples) and McNemar's test. The both-correct subset from which these 100 rows are drawn constitutes 38.2\% of VQAv2 and 90.0\% of COCO Captions; see \S\ref{sec:negation_results} for full subset statistics.}

\begin{table*}[t]
\centering
\caption{Gaussian blur robustness (kernel $5{\times}5$, $\sigma{=}2.0$, $n{=}100$ both-correct). Drop shown with 95\% bootstrap CI (10{,}000 resamples, percentile method). Original accuracy is 100\% by construction. McNemar $p$ tests the paired null of no differential degradation.}
\label{tab:blur}
\small
\begin{tabular*}{\textwidth}{@{\extracolsep{\fill}}lcccc@{}}
\toprule
\textbf{Model} & \textbf{Clean} & \textbf{Blurred} & \textbf{Drop} & \textbf{95\% CI} \\
\midrule
\smol{} (FP16) & 100.0\% & 97.0\% & 3.0\pp & [0.0, 7.0]\pp \\
\qwen{} (NF4) & 100.0\% & 97.0\% & 3.0\pp & [0.0, 7.0]\pp \\
\midrule
\multicolumn{4}{@{}l}{Sensitivity ratio $\rho$} & 1.00~[0.00, 5.00] \\
\multicolumn{4}{@{}l}{McNemar $p$ (paired, continuity-corrected)} & 0.683 \\
\bottomrule
\end{tabular*}
\end{table*}

The bootstrap CIs span zero for both models, and McNemar's test ($p = 0.683$, 3 discordant pairs each) detects no significant differential degradation at $n{=}100$. Two controlled ablations (Table~\ref{tab:blur_ablation}) examine whether quantisation or scale modulate this equal sensitivity.

\begin{table*}[t]
\centering
\caption{Four-model blur ablation (kernel $5{\times}5$, $\sigma{=}2.0$, $n{=}100$). Original accuracy for SmolVLM2 and Qwen-7B NF4 is 100\% by construction; Qwen-7B FP16 baseline is 97\% and Qwen-3B baseline is 81\% (these models were not part of the selection criterion).}
\label{tab:blur_ablation}
\small
\begin{tabular}{@{}llccccc@{}}
\toprule
\textbf{Model} & \textbf{Prec.} & \textbf{Clean} & \textbf{Blurred} & \textbf{Drop} & \textbf{Ret.}$^\dagger$ \\
\midrule
\smol{} (0.5B) & FP16 & 100.0\% & 97.0\% & $-$3.0\pp & 0.970 \\
Qwen2.5-VL-3B$^\ddagger$ & FP16 & 81.0\% & 77.0\% & $-$4.0\pp & 0.951 \\
\qwen{} & FP16 & 97.0\% & 97.0\% & $\phantom{+}$0.0\pp & 1.000 \\
\qwen{} & NF4 & 100.0\% & 97.0\% & $-$3.0\pp & 0.970 \\
\midrule
\multicolumn{5}{@{}l}{$\rho_{\text{orig}}$ (SmolVLM2 / Qwen-7B NF4)} & 1.00 \\
\multicolumn{5}{@{}l}{$\rho_{\text{precision}}$ (SmolVLM2 / Qwen-7B FP16)} & $\infty$ (degenerate: 0\pp denom.) \\
\multicolumn{5}{@{}l}{$\rho_{\text{scale}}$ (Qwen-3B / Qwen-7B, both FP16)} & $\infty$ (degenerate: 0\pp denom.) \\
\multicolumn{5}{@{}l}{$\rho_{\text{arch}}$ (SmolVLM2 / Qwen-3B, both FP16)} & \textbf{0.75} \\
\bottomrule
\multicolumn{6}{@{}l}{$^\dagger$Retention = blurred/clean (own baseline). $^\ddagger$Qwen-3B baseline is 81\%: the}  \\
\multicolumn{6}{@{}l}{100-image pool was drawn from rows correct for SmolVLM2 \& Qwen-7B NF4, not Qwen-3B.} \\
\end{tabular}
\end{table*}

\begin{figure}[t]
\centering
\includegraphics[width=0.9\linewidth]{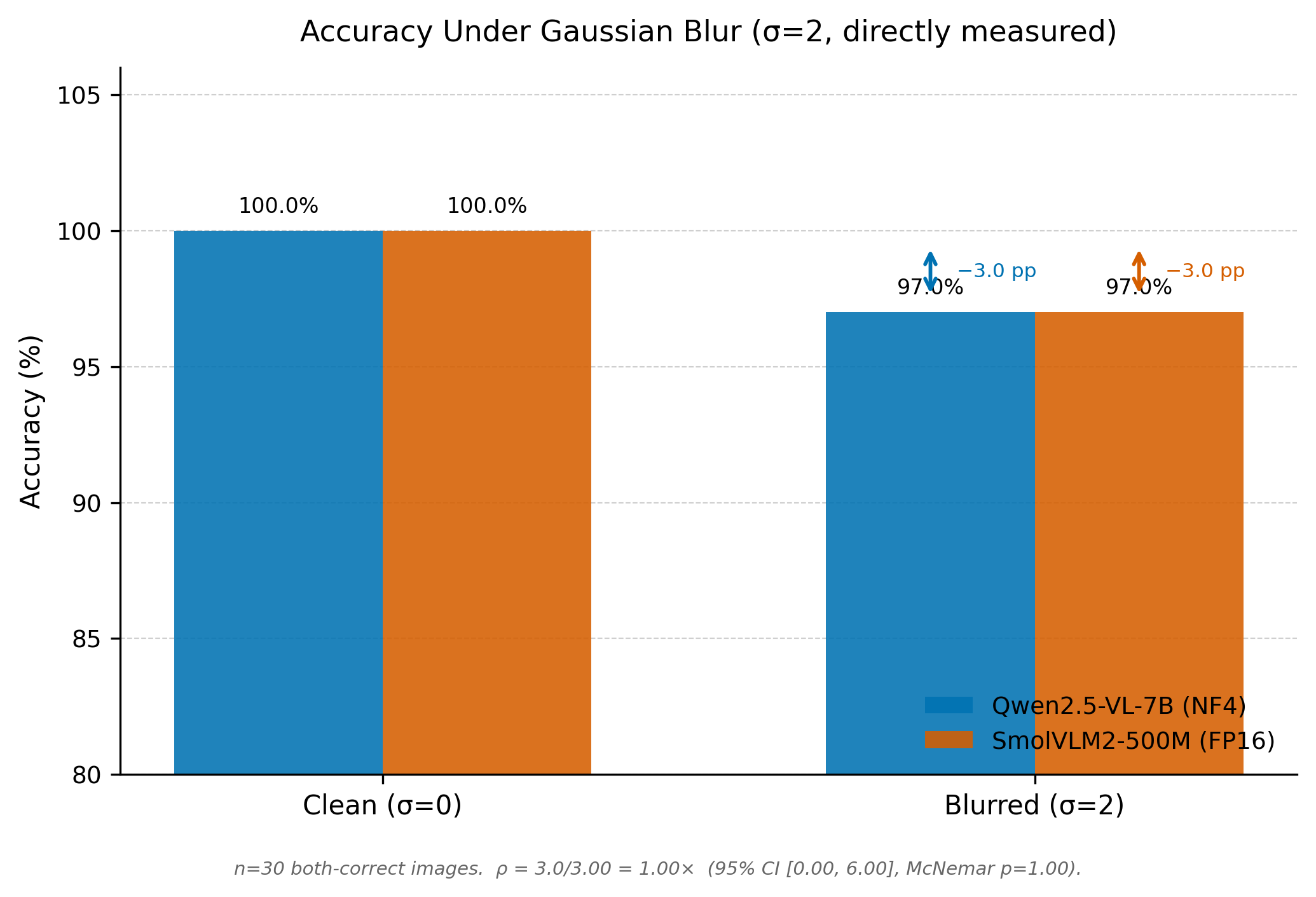}
\caption{Accuracy under clean ($\sigma{=}0$) and Gaussian-blurred ($\sigma{=}2$) conditions for both models ($n{=}100$ both-correct images). Both conditions are directly measured; no extrapolation is performed. Both models drop $3.0$\pp, giving $\rho = 1.00$ (95\% CI $[0.00, 5.00]$; McNemar $p = 0.683$; differential degradation not significant).}
\label{fig:robustness}
\end{figure}

Three findings emerge from the $n{=}100$ ablation. First, NF4 quantisation adds blur sensitivity: the NF4 variant drops $3.0\pp$ whereas the FP16 variant shows $0\pp$ drop (97\%$\to$97\%), so the precision ratio is degenerate ($\rho_{\text{precision}} = \infty$). Second, scale within the Qwen architecture confers robustness: Qwen-7B FP16 ($0\pp$ drop) is more robust than Qwen-3B FP16 ($4\pp$ drop), also with a degenerate ratio ($\rho_{\text{scale}} = \infty$). Third, at comparable precision and scale, SmolVLM2 ($3\pp$) is \emph{marginally less} sensitive than Qwen-3B ($4\pp$), giving $\rho_{\text{arch}} = 0.75$---counter to the SigLIP encoder bottleneck hypothesis. The degenerate $\rho$ values for precision and scale indicate that Qwen-7B FP16 is fully robust to this blur level on this pool; the $4\pp$ Qwen-3B drop should be interpreted cautiously given its lower baseline accuracy (81\%) on the SmolVLM2-selected pool. Figure~\ref{fig:robustness} shows the clean vs.\ blurred accuracy for both models at $\sigma{=}2$.

\textbf{Interpreting degenerate $\rho$ values.} Two entries show $\rho = \infty$ because Qwen-7B~FP16 shows $0\pp$ degradation (97\%$\to$97\%), making the denominator zero. The key signal is directional: SmolVLM2 and Qwen-NF4 both degrade by $3\pp$, Qwen-3B by $4\pp$, while Qwen-7B~FP16 is fully robust at this blur level.

\textbf{Both-correct subset caveat.} Sampling only from rows where both models were originally correct ensures a clean 100\% baseline but may over-represent visually simple images, limiting headroom for degradation. A companion script (\texttt{stratified\_robustness.py}) extends the experiment to all four correctness strata; results are left as future work.

\subsection{Negation Stress Tests}
\label{sec:negation_results}

Table~\ref{tab:negation} presents per-dataset negation accuracy. \textbf{The negation gap is strongly dataset-dependent}: on COCO, \smol{} degrades $20.5\pp$ more than \qwen{} (95\% CI $[15.8, 25.2]$, $p < 10^{-16}$), while on VQAv2 the gap is only $4.5\pp$ and \textbf{not statistically significant} ($p{=}0.19$). The aggregate gap---$12.5\pp$ (95\% CI $[8.2, 16.8]\pp$, $z{=}5.69$, $p < 10^{-8}$)---is therefore driven almost entirely by the COCO templates.
\textbf{Instruction-following asymmetry caveat.} The \texttt{false\_yn} template (``Is it true that [answer] is NOT shown in this image?'') uses the model's own correct answer and requires rejection of a false negation. This is a well-designed logical inversion probe, but it also tests instruction-following capability: \qwen{}'s 7B instruction-tuned backbone may find it inherently easier to parse and comply with the complex prompt structure, independent of negation reasoning \emph{per se}. The 0\% result for \smol{} is therefore partly confounded with the model's general instruction-following capacity. A control template testing other forms of logical inversion---not involving the model's own prior answer---would help isolate negation reasoning specifically; this is left as future work.

\begin{table*}[t]
\centering
\caption{Aggregate negation accuracy ($n{=}100$ both-correct rows per dataset {\texttimes} 4 templates = 800 judgements per model). Baseline accuracy is 100\% by construction. Wilson 95\% CIs in brackets.}
\label{tab:negation}
\small
\begin{tabular}{@{}lcccc@{}}
\toprule
\textbf{Model} & \textbf{Baseline} & \textbf{Negation} & \textbf{95\% CI} & \textbf{Drop} \\
\midrule
\smol{} (FP16) & 100.0\% & 66.8\% & [63.4, 69.9] & $-$33.2\pp \\
\qwen{} (NF4) & 100.0\% & 79.2\% & [76.3, 81.9] & $-$20.8\pp \\
\midrule
\multicolumn{3}{@{}l}{\textbf{Gap (Qwen $-$ Smol)}} & & 12.5\pp \\
\multicolumn{3}{@{}l}{\quad Wald 95\% CI} & & [8.2, 16.8]\pp \\
\multicolumn{3}{@{}l}{\quad $z = 5.69$, $p < 10^{-8}$} & & \\
\bottomrule
\end{tabular}
\end{table*}

\textbf{Per-template breakdown.} Table~\ref{tab:negation_detail} reveals highly template-specific behaviour. The most discriminating template is \texttt{false\_yn}: \smol{} achieves 2\% success on VQAv2 (Wilson 95\% CI [0.6, 7.0]) and \textbf{0\% on COCO} (CI [0.0, 3.7])---responding ``Yes'' (the hallucination response) on 98--100\% of trials. \qwen{} scores 49\% (CI [39.4, 58.7]) and 86\% (CI [77.9, 91.5]) respectively. The \texttt{counter} template shows a similar asymmetry on VQAv2 (39\% vs.\ 100\%). Conversely, both models handle \texttt{is\_not} and \texttt{absent} well on COCO ($\geq$97\%), suggesting that open-ended negation is easier than logical inversion under a false premise.

\textbf{Per-dataset breakdown.} The aggregate gap masks a dramatic asymmetry: COCO accounts for nearly all of it ($20.5\pp$, $p < 10^{-16}$), driven by the \texttt{false\_yn} collapse, while VQAv2 is not significant ($4.5\pp$, $p{=}0.19$). The strongest negation claim rests entirely on COCO.
\begin{table*}[t]
\centering
\caption{Per-template negation success rates (\%) across VQAv2 and COCO. $n{=}100$ per cell. Bold indicates the most discriminating comparisons.}
\label{tab:negation_detail}
\small
\setlength{\tabcolsep}{3pt}
\begin{tabular}{@{}l cc cc@{}}
\toprule
& \multicolumn{2}{c}{\textbf{VQAv2}} & \multicolumn{2}{c}{\textbf{COCO}} \\
\cmidrule(lr){2-3} \cmidrule(lr){4-5}
\textbf{Template} & \smol & \qwen & \smol & \qwen \\
\midrule
\texttt{is\_not} & 94 & 47 & 100 & 97 \\
\texttt{absent} & 99 & 56 & 100 & 99 \\
\texttt{false\_yn} & \textbf{2} & \textbf{49} & \textbf{0} & \textbf{86} \\
\texttt{counter} & \textbf{39} & \textbf{100} & 100 & 100 \\
\midrule
Average & 58.5 & 63.0 & 75.0 & 95.5 \\
\bottomrule
\end{tabular}
\end{table*}

\begin{figure*}[t]
\centering
\includegraphics[width=\textwidth]{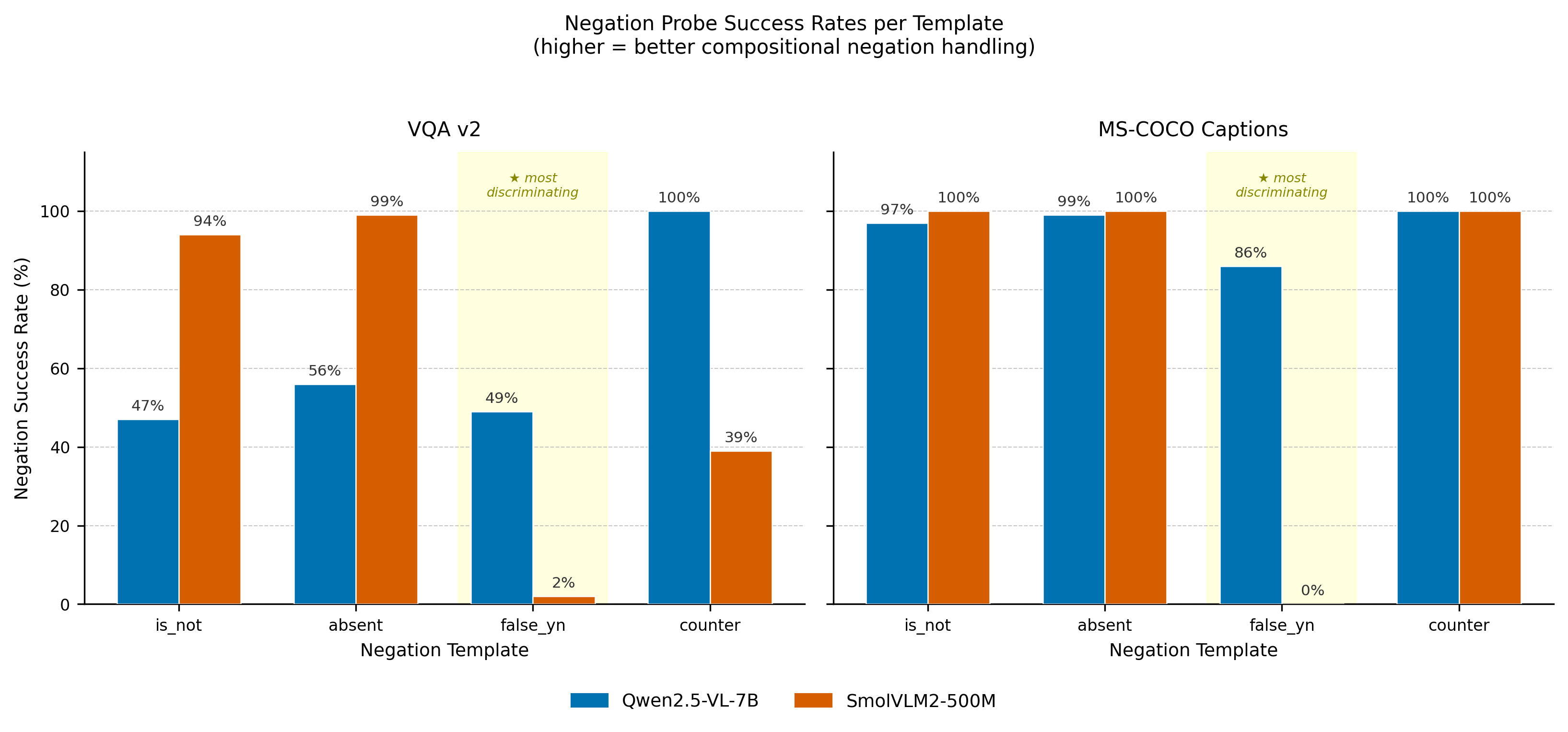}
\caption{Negation probe success rates per template across VQAv2 (left) and COCO Captions (right). \qwen{} (blue) and \smol{} (red) are shown side-by-side. The \texttt{false\_yn} template is the most discriminating: \smol{} collapses to 2\% on VQAv2 and 0\% on COCO, while \qwen{} scores 49\% and 86\%.}
\label{fig:negation}
\end{figure*}

\textbf{Interpretation.} The \texttt{false\_yn} collapse suggests \smol{} processes the negated prompt as a positive object-identification question, responding ``Yes'' as if ``NOT'' were absent---achieving 0\% on COCO. This failure mode is invisible to standard accuracy metrics. Interestingly, \smol{} \emph{outperforms} \qwen{} on \texttt{is\_not} and \texttt{absent} for VQAv2 (94/99\% vs.\ 47/56\%), possibly because \qwen{}'s larger language model anchors more strongly to the original answer---though tokenisation and instruction-tuning differences could also contribute. This cross-template divergence shows that negation comprises distinct compositional skills, not a monolithic capability. Figure~\ref{fig:negation} shows the full breakdown.

\textbf{Both-correct subset caveat.} The $12.5\pp$ negation gap is measured only on both-correct images (38.2\% of VQAv2, 90.0\% of COCO). This set is enriched for visually unambiguous cases; the negation gap on harder images is uncharacterised. For COCO, 90\% coverage makes the subset nearly representative. The \texttt{false\_yn} collapse is therefore a pure compositional failure; the total deficit on arbitrary images remains uncharacterised.

Under the most pessimistic assumption (zero additional collapse outside the both-correct set), selection-weighted lower bounds are ${\approx}1.7\pp$ for VQAv2 and ${\approx}18.5\pp$ for COCO. The VQAv2 bound is notably small, confirming selection bias is a material concern there. However, harder images likely exhibit \emph{greater} negation collapse, so true population-level gaps may exceed the measured values. The \texttt{false\_yn} finding (0\% for \smol{} on COCO) is unaffected by selection bias, as it operates only on correctly-identified objects.

\section{Discussion}
\label{sec:discussion}

\subsection{Structural Characterisation of the Reliability Gap}

For this specific model pair, the compact model fails differently, not just more often. The four evaluation axes---taxonomy, calibration, blur robustness, and negation---reveal qualitatively distinct failure profiles. Because the two models differ simultaneously in scale, architecture, and numerical precision (\S\ref{sec:confound}), the interpretations below are \emph{candidate explanations} consistent with the observed patterns, not causally established conclusions.

\textbf{(1) No differential blur sensitivity; quantisation and scale effects directional.} No significant blur gap is detected (Table~\ref{tab:blur}). The controlled ablations (Table~\ref{tab:blur_ablation}) reveal directional signals: NF4 quantisation adds sensitivity, scale confers robustness, and the SigLIP encoder bottleneck hypothesis is not supported ($\rho_{\text{arch}} = 0.75$). These remain directional at $n{=}100$.

\textbf{(2) Language head dominance (hypothesised).} The \texttt{false\_yn} collapse (0--2\% success) shows \smol{} affirming false premises about present objects being absent on 98--100\% of trials---consistent with its 135M-parameter language backbone overriding visual reasoning. This does not constitute mechanistic proof: the collapse could also arise from instruction-tuning differences, tokenisation artefacts, or negation-specific training data coverage. The template also tests instruction-following capacity; \qwen{}'s larger backbone may parse the complex prompt more easily, independent of negation ability (\S\ref{sec:negation_results}).

\textbf{(3) Calibration as a deployment signal (operational, not diagnostic).} The asymmetric calibration pattern (Table~\ref{tab:ece}, Figure~\ref{fig:reliability}) means confidence scores carry different informativeness depending on task and model. Regardless of whether \qwen{}'s constant $p\approx 0.999$ on VQAv2 reflects genuine calibration collapse or a tokenisation artefact, the practical consequence is the same: zero discriminative value for deployment gating. No single threshold policy works across tasks for either model. The ECE differences are a joint function of scale and precision (NF4 vs.\ FP16) and cannot be attributed to either factor alone.

\subsection{The Quantisation Confound}
\label{sec:confound}

A critical limitation is that \qwen{} runs in 4-bit NF4 while \smol{} runs in FP16. The observed reliability gap is therefore a joint function of: (i)~model scale (7B vs.\ 0.5B), (ii)~numerical precision (NF4 vs.\ FP16), and (iii)~architecture (Qwen2.5-VL vs.\ SmolVLM2). Recent work on quantisation-specific reliability effects~\cite{bouguerra2025quantclip,bhatnagar2025luq,slyman2024quantbias} suggests that 4-bit quantisation can \emph{either} improve or worsen calibration depending on whether the base model is under- or over-confident, making the ECE comparison a joint effect rather than a pure scale effect.

\textbf{FP16 precision ablation (completed, $n{=}100$).} \qwen{} in native FP16 was re-run on the same 100 both-correct blur rows as \S\ref{sec:robustness} using (hardware: RTX 5090, 32\,GiB VRAM; footprint $\approx$16\,GiB). Qwen-7B FP16 shows $0\pp$ drop (97\%$\to$97\%), while Qwen-7B NF4 drops $3\pp$ (100\%$\to$97\%). Note that Qwen-7B FP16 starts at 97\% original accuracy on this pool (not 100\%), because the pool was selected for SmolVLM2 / Qwen-NF4 correctness. The $3\pp$ precision effect (NF4 vs.\ FP16) directionally confirms that quantisation hurts blur robustness; $\rho_{\text{precision}}$ is degenerate ($\infty$) because the FP16 denominator is $0\pp$.

\textbf{Scale ablation within Qwen architecture (completed, $n{=}100$).} Qwen2.5-VL-3B in FP16 was run on the same 100 rows using (VRAM: $\approx$7\,GiB). Qwen-3B drops $4.0\pp$ (81\%$\to$77\%), while \smol{} drops $3.0\pp$, giving $\rho_{\text{arch}} = 3.0/4.0 = 0.75$ (Table~\ref{tab:blur_ablation}). SmolVLM2 is \emph{marginally less} blur-sensitive than Qwen-3B at comparable precision, counter to the SigLIP encoder bottleneck hypothesis. Scale within the Qwen architecture shows a clear benefit: Qwen-7B FP16 ($0\pp$) is substantially more robust than Qwen-3B FP16 ($4\pp$), consistent with the larger LLM backbone reconstructing semantics from degraded visual tokens, though the $\rho_{\text{scale}}$ ratio is degenerate ($\infty$). The Qwen-3B baseline on this pool is only 81\% (because the pool was selected for correctness on SmolVLM2 and Qwen-7B, not Qwen-3B), so the $4\pp$ drop estimate carries a low-baseline caveat.

\subsection{Error Taxonomy Interpretation}

Semantic Drift~(B) dominates VQAv2 failures for both models (Table~\ref{tab:taxonomy}), consistent with correct object identification but wrong attributes. \qwen{} shows notably more Prior Bias~(C) on VQAv2 (20.4\% vs.\ 4.2\%), suggesting its stronger language backbone generates more plausible-but-wrong predictions anchored to scene priors. The failure-profile concordance ($\kappa = 0.017$ on VQAv2, $-0.038$ on COCO) confirms divergent failure structures, particularly on COCO where \qwen{} concentrates on Semantic Drift while \smol{} splits between Object Blindness and Semantic Drift.

The GPT-4o judge operates text-only and cannot verify object presence in images; a vision-capable judge would strengthen COCO categorisation.

\subsection{Implications for Edge Deployment}

These findings have direct implications for edge deployment:

\textbf{Safety-critical applications.} The \texttt{false\_yn} collapse in \smol{} is particularly concerning for safety-critical applications. The template asks ``Is it true that [depicted object] is NOT shown?''---the correct answer is ``No''. \smol{} answers ``Yes'' on 98--100\% of trials, effectively agreeing that present objects are absent. In an operational context, a model that confirms ``Is it true that no obstacles are present?'' with ``Yes'' when obstacles exist represents a dangerous false safety assertion, not merely a negation reasoning lapse. Models deployed in autonomous driving, medical imaging, or security contexts must be stress-tested with such logical-inversion probes before deployment.

\textbf{Confidence-based gating.} Confidence thresholds for routing uncertain queries collapse for \qwen{} on VQAv2: constant $\approx$0.999 confidence admits every prediction, including the 44.4\% that are wrong. For \smol{} on COCO, underconfidence over-routes correct answers to fallbacks. Task-adaptive calibration~\cite{guo2017calibration} is essential; off-the-shelf confidence gates without per-task recalibration risk silent failures.

\textbf{Multi-modal robustness testing.} A minimum deployment-readiness suite should include: (i)~blur/noise perturbation at ImageNet-C severity levels 1--3; (ii)~negation probes across at least two templates; and (iii)~ECE measurement with reliability diagrams. The released pipeline provides a reference implementation.

\section{Conclusion}
\label{sec:conclusion}

The central finding is qualitative: \textbf{for this model pair, the compact model fails differently, not just more often}. The error taxonomy reveals distinct failure profiles (\S\ref{sec:taxonomy_results}); negation probes expose a striking compositional collapse in \smol{}, driven by COCO (\S\ref{sec:negation_results}); calibration analysis uncovers asymmetric, dataset-dependent miscalibration that renders confidence-based gating ineffective (\S\ref{sec:ece_results}); and blur ablations provide directional evidence on quantisation and scale effects (\S\ref{sec:robustness}). As a single model pair ($N{=}1$), generalisation is the most important open question.

Aggregate accuracy is insufficient for edge deployment readiness. Compressed VLMs should undergo structured reliability auditing---calibration analysis, perturbation stress-testing, and compositional reasoning probes---before safety-relevant deployment. The taxonomy and pipeline introduced here provide a reusable framework; all code and artefacts are released for full reproducibility.

\clearpage
\onecolumn
\appendix
\section*{Supplementary Material}
\setcounter{section}{0}
\renewcommand{\thesection}{\Alph{section}}

\section{Detailed Calibration Bin Statistics}
\label{app:calibration}

Tables~\ref{tab:ece_smol_vqa}--\ref{tab:ece_qwen_coco} report the full $M{=}10$ bin statistics for all model--dataset combinations. The contrast between \smol{}'s distributed confidence on VQAv2 (predictions spread across bins 2--9) and \qwen{}'s extreme concentration (all 2{,}000 predictions in bin 9) is the most salient feature.

\begin{table}[ht]
\centering
\caption{\smol{} calibration bins on VQAv2 ($n{=}2{,}000$, ECE $= 0.228$).}
\label{tab:ece_smol_vqa}
\small
\begin{tabular}{@{}lccc@{}}
\toprule
\textbf{Bin range} & \textbf{Count} & \textbf{Avg.\ Conf.} & \textbf{Accuracy} \\
\midrule
$[0.2, 0.3)$ & 1 & 0.268 & 0.0\% \\
$[0.3, 0.4)$ & 23 & 0.373 & 39.1\% \\
$[0.4, 0.5)$ & 148 & 0.459 & 46.0\% \\
$[0.5, 0.6)$ & 305 & 0.554 & 42.3\% \\
$[0.6, 0.7)$ & 448 & 0.655 & 45.3\% \\
$[0.7, 0.8)$ & 529 & 0.751 & 45.8\% \\
$[0.8, 0.9)$ & 402 & 0.843 & 49.8\% \\
$[0.9, 1.0)$ & 144 & 0.933 & 72.2\% \\
\bottomrule
\end{tabular}
\end{table}

\begin{table}[ht]
\centering
\caption{\smol{} calibration bins on COCO ($n{=}2{,}000$, ECE $= 0.431$).}
\label{tab:ece_smol_coco}
\small
\begin{tabular}{@{}lccc@{}}
\toprule
\textbf{Bin range} & \textbf{Count} & \textbf{Avg.\ Conf.} & \textbf{Accuracy} \\
\midrule
$[0.2, 0.3)$ & 18 & 0.286 & 77.8\% \\
$[0.3, 0.4)$ & 223 & 0.370 & 90.1\% \\
$[0.4, 0.5)$ & 867 & 0.456 & 91.4\% \\
$[0.5, 0.6)$ & 713 & 0.543 & 94.3\% \\
$[0.6, 0.7)$ & 170 & 0.633 & 93.5\% \\
$[0.7, 0.8)$ & 8 & 0.730 & 87.5\% \\
$[0.8, 0.9)$ & 1 & 0.818 & 0.0\% \\
\bottomrule
\end{tabular}
\end{table}

\begin{table}[ht]
\centering
\caption{\qwen{} calibration bins on VQAv2 ($n{=}2{,}000$, ECE $= 0.443$). All predictions fall in a single bin, indicating extreme overconfidence.}
\label{tab:ece_qwen_vqa}
\small
\begin{tabular}{@{}lccc@{}}
\toprule
\textbf{Bin range} & \textbf{Count} & \textbf{Avg.\ Conf.} & \textbf{Accuracy} \\
\midrule
$[0.9, 1.0)$ & 2{,}000 & 0.999 & 55.6\% \\
\emph{All other bins} & 0 & --- & --- \\
\bottomrule
\end{tabular}
\end{table}

\begin{table}[ht]
\centering
\caption{\qwen{} calibration bins on COCO ($n{=}2{,}000$, ECE $= 0.087$). All predictions fall in a single bin; high task accuracy keeps ECE low.}
\label{tab:ece_qwen_coco}
\small
\begin{tabular}{@{}lccc@{}}
\toprule
\textbf{Bin range} & \textbf{Count} & \textbf{Avg.\ Conf.} & \textbf{Accuracy} \\
\midrule
$[0.9, 1.0)$ & 2{,}000 & 0.998 & 91.1\% \\
\emph{All other bins} & 0 & --- & --- \\
\bottomrule
\end{tabular}
\end{table}

\section{Qualitative Failure Examples}
\label{app:qualitative}

\textbf{Example 1 --- Category B (Semantic Drift), VQAv2.} [Manually verified; heuristic assigns A due to yes/no GT structure.]
\emph{Q:} ``Are the boats seaworthy?'' \emph{GT:} ``yes.''
\smol{}: ``The boats are not seaworthy, as they are not designed for such conditions.''
\qwen{}: ``It is not possible to definitively determine from the image alone whether the boat in the foreground is seaworthy\ldots''
By visual inspection, both models identify the boats (correct object class) but produce wrong attribute judgements about seaworthiness---a Semantic Drift failure. The heuristic incorrectly assigns Category~A because the GT token ``yes'' does not appear in predictions (zero text overlap). This is a concrete illustration of the heuristic's failure mode on yes/no VQAv2 questions: the A-vs-B distinction collapses to whether the GT string happens to appear in the output, not whether the model identified the correct object.

\textbf{Example 2 --- Category A (Object Blindness), COCO.}
\emph{Prompt:} ``Generate a short caption for this image.'' \emph{GT:} ``Two women waiting at a bench next to a street.''
\smol{}: ``A woman is sitting on a bench and talking to another woman.''
\qwen{}: ``Two individuals wait at a bus stop in an urban setting with modern buildings and palm trees in the background.''
Heuristic classifier: Both models miss or under-specify primary scene elements.

\textbf{Example 3 --- Category D (Spatial Error), VQAv2.}
\emph{Q:} ``How many wheels do you see?'' \emph{GT:} ``6.''
\smol{}: ``There are two wheels visible in the image.''
\qwen{}: ``In the image, there are two wheels visible on the motorcycle in the foreground.''
Heuristic classifier: Both models undercount visible wheels --- a spatial/quantitative reasoning error.

\section{Negation Probe Template Prompts}
\label{app:negation_prompts}

\begin{table}[ht]
\centering
\caption{Full negation probe template prompts. \texttt{\{answer\}} is replaced with the model's original correct answer; \texttt{\{question\}} with the original question text.}
\small
\begin{tabular}{@{}lp{10cm}@{}}
\toprule
\textbf{Template} & \textbf{Prompt text} \\
\midrule
\texttt{is\_not} & ``What object or thing is NOT depicted in this image?'' \\
\texttt{absent} & ``Name something absent from this image.'' \\
\texttt{false\_yn} & ``Is it true that `\{answer\}' is NOT shown in this image? Answer yes or no.'' \\
\texttt{counter} & ``The answer to the question `\{question\}' is NOT `\{answer\}'. What is the correct answer?'' \\
\bottomrule
\end{tabular}
\end{table}

\section{GPT-4o Judge Prompt}
\label{app:judge_prompt}

\begin{verbatim}
System: You are an expert evaluator of vision-language
model outputs. Classify the following model error into
exactly one category:
[A: Object Blindness | B: Semantic Drift | C: Prior Bias |
 D: Spatial/Relational Error | E: Other].
Definitions:
  A - Model fails to identify a salient visible object.
  B - Correct object category, wrong attribute
      (colour, count, action).
  C - Plausible output consistent with scene priors,
      wrong for this image.
  D - Incorrect spatial relation or layout description.
  E - Does not clearly fit A-D.

User:   Dataset: {dataset}
        Question/Prompt: {question}
        Ground Truth: {ground_truth}
        Model Output: {prediction}
        Respond with JSON:
        {"category":"A|B|C|D|E",
         "confidence": 0-1,
         "reasoning": "..."}
\end{verbatim}

\section{Reproducibility Details}
\label{app:reproducibility}

All experiments share a deterministic seed (\texttt{SEED=42}) applied to Python \texttt{random}, \texttt{os.environ["PYTHONHASHSEED"]}, \texttt{torch.manual\_seed}, and \texttt{torch.cuda.manual\_seed\_all}. CuDNN is run in deterministic mode with benchmarking disabled. The 100-sample robustness subset is selected with \texttt{random.sample(both\_correct, 100)} after this seed, ensuring identical subsets across runs given the same inference CSV. The complete pipeline can be executed via a single command:

\begin{verbatim}
bash run_pipeline.sh
\end{verbatim}

\noindent All phases auto-commit their outputs to version control. Hardware: NVIDIA RTX 5090 (32\,GiB VRAM), single-GPU. Software: PyTorch 2.x, \texttt{transformers} 4.x, \texttt{bitsandbytes}, \texttt{datasets} (streaming mode).

\end{document}